\pdfoutput=1

\documentclass[11pt]{article}

\usepackage{acl}

\usepackage{times}
\usepackage{latexsym}

\usepackage[T1]{fontenc}

\usepackage[utf8]{inputenc}

\usepackage{microtype}

\usepackage{inconsolata}

\usepackage{xcolor}
\usepackage{ragged2e}
\usepackage{enumitem}
\usepackage{graphicx}
\usepackage{booktabs}
\usepackage{amsmath,amssymb,amsfonts}
\usepackage{algorithm}
\usepackage[noend]{algpseudocode}
\usepackage{algorithmicx}
\algrenewcommand\algorithmicrequire{\textbf{Input:}}
\algrenewcommand\algorithmicensure{\textbf{Output:}}

\title{Few-Shot Dialogue Summarization via Skeleton-Assisted Prompt Transfer in Prompt Tuning}

\author{Kaige Xie\textsuperscript{$\clubsuit$} \quad Tong Yu\textsuperscript{$\dagger$} \quad Haoliang Wang\textsuperscript{$\dagger$}\Thanks{\ Corresponding author} \quad Junda Wu\textsuperscript{$\diamondsuit$} \\ \bf Handong Zhao\textsuperscript{$\dagger$} \quad Ruiyi Zhang\textsuperscript{$\dagger$} \quad Kanak Mahadik\textsuperscript{$\dagger$} \quad Ani Nenkova\textsuperscript{$\dagger$} \quad Mark Riedl\textsuperscript{$\clubsuit$} \\
\textsuperscript{$\clubsuit$} School of Interactive Computing, Georgia Institute of Technology \\
\textsuperscript{$\dagger$}Adobe Research \qquad \textsuperscript{$\diamondsuit$}University of California San Diego \\
\texttt{\{kaigexie, riedl\}@gatech.edu} \qquad \texttt{juw069@ucsd.edu}\\
\texttt{\{tyu, hawang, hazhao, ruizhang, mahadik, nenkova\}@adobe.com}\\}

\begin{document}
\maketitle
\begin{abstract}
In real-world scenarios, labeled samples for dialogue summarization are usually limited (i.e., few-shot) due to high annotation costs for high-quality dialogue summaries.
To efficiently learn from few-shot samples, previous works have utilized massive annotated data from other downstream tasks and then performed prompt transfer in prompt tuning so as to enable cross-task knowledge transfer.
However, existing general-purpose prompt transfer techniques lack consideration for dialogue-specific information.
In this paper, we focus on improving the prompt transfer from dialogue state tracking to dialogue summarization and propose Skeleton-Assisted Prompt Transfer (SAPT), which leverages skeleton generation as extra supervision that functions as a medium connecting the distinct source and target task and resulting in the model's better consumption of dialogue state information.
To automatically extract dialogue skeletons as supervised training data for skeleton generation, we design a novel approach with perturbation-based probes requiring neither annotation effort nor domain knowledge.
Training the model on such skeletons can also help preserve model capability during prompt transfer.
Our method significantly outperforms existing baselines.
In-depth analyses demonstrate the effectiveness of our method in facilitating cross-task knowledge transfer in few-shot dialogue summarization.
\end{abstract}

\begin{figure}[ht!]
\centering
\includegraphics[width=0.85\linewidth]{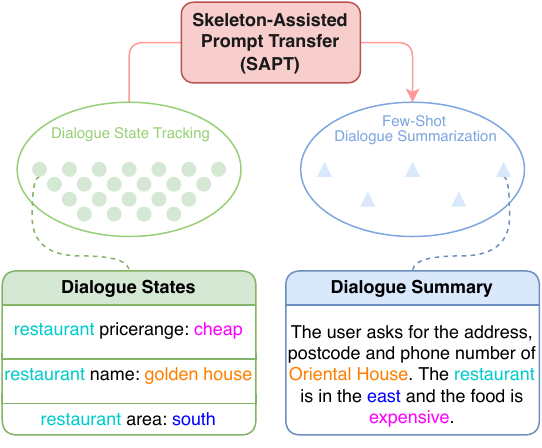}
\caption{We study the problem of how to perform effective transfer learning from dialogue state tracking (DST) to few-shot dialogue summarization, in the scenario where there is a large set of dialogues with DST annotations, and another small set of dialogues with dialogue summarization annotations (i.e., few-shot learning for dialogue summarization).}
\label{fig:intro}
\end{figure}

\section{Introduction} 

Automatic text summarization \cite{luhn1958automatic} is one of the most important and challenging problems in NLP. Among the different forms the text to be summarized could take, dialogues have been serving as a critical part of human-human and human-machine interaction. There has been significant progress made in dialogue summarization these days \cite{goo2018abstractive, liu2019topic, chen-yang-2020-multi}.
However, they generally rely on massive human-written golden dialogue summaries.
In real-world scenarios, the availability of massive supervised data is not always guaranteed, as the data scarcity problem often occurs due to the high annotation cost that is normally required for acquiring large-scale high-quality dialogue summaries \cite{brazinskas-etal-2020-shot}.

In existing works, one common way to tackle the data scarcity problem is to perform transfer learning by leveraging off-the-shelf out-of-domain or out-of-task supervised data \cite{yang-etal-2020-ted,goodwin-etal-2020-towards,yu-etal-2021-adaptsum,zou-etal-2021-low,magooda-etal-2021-exploring-multitask}.
We observe that the supervised data of a relevant task called dialogue state tracking (DST) \cite{williams2007partially} can bring conducive knowledge for the dialogue summarization task, as the semantic slots and values tracked by DST are expected to be covered in the dialogue summary \cite{shin-etal-2022-dialogue}. Besides the notable relevance between those two tasks, with DST being a language understanding task as opposed to dialogue summarization being a language generation task, the annotations of DST should arguably be easier to get in practice than those of dialogue summarization.\footnote{In \autoref{sec:data-annotation-appendix}, we validate it via a data annotation study.} These observations motivate us to herein focus on developing effective transfer learning techniques for the scenario where there are ample supervised data for DST whereas the annotations for dialogue summarization are limited, as depicted in \autoref{fig:intro}.

Among recent transfer learning techniques, prompt transfer \cite{vu-etal-2022-spot} in prompt tuning \cite{li-liang-2021-prefix,lester-etal-2021-power} has gained great popularity because of its parameter efficiency. Prompt tuning is a paradigm of utilizing pretrained language models (PLMs) for downstream tasks, in which a sequence of continuous trainable embeddings called ``soft prompt'' is prepended to the input sequence so as to provide PLMs with an adequate context. During training, only these embeddings can be updated while all the other parameters of PLMs will remain fixed. Prompt transfer realizes cross-task transfer learning under the prompt tuning paradigm by training soft prompts from source tasks and then using them as parameter initialization for the prompt tuning in target tasks. In general, prompt transfer works well in transfer learning between language understanding tasks while it can only provide relatively mediocre performance in language generation tasks \cite{su-etal-2022-transferability}, indicating \emph{the necessity to design task-specific prompt transfer approaches} for language generation tasks such as dialogue summarization.

How to improve prompt transfer \emph{in a task-specific manner}? The existing general-purpose prompt transfer technique \cite{vu-etal-2022-spot} relies solely on the source and target task supervision, suggesting the lack of an intermediate task-specific medium that could potentially better connect the distinct source and target task. Also, as the model capability of processing source task data is closely associated with the knowledge it has gained during the source task pretraining, it needs to be effectively preserved during the prompt transfer so as to facilitate the model in handling the target task.

In this paper, we propose a dialogue-specific prompt transfer technique, named \textbf{S}keleton-\textbf{A}ssisted \textbf{P}rompt \textbf{T}ransfer (SAPT).
SAPT provides the model with extra supervision during its prompt transfer by training it to perform skeleton generation along the way. This extra supervision can essentially function as an \emph{intermediate task-specific medium} that is beneficial for the knowledge transfer between the distinct source and target task. To get the supervised training data for skeleton generation, we design a novel automatic skeleton extraction approach that requires neither annotation effort nor domain knowledge. Specifically, we observe the model's output variation to perturbation-based probes and extract the dialogue turns to which the model displays the highest sensitivity as skeletons. Training the model on such skeletons can also help \emph{preserve model capability during prompt transfer}. The idea behind this is that we try to prevent the model from forgetting the dialogue-state-related knowledge it has learned during its pretraining on supervised DST data, since the model sensitivity to perturbation-based probes in the DST task intrinsically reflects the capability of processing dialogue state information it has developed. Experimental results and in-depth analyses with BART \cite{lewis-etal-2020-bart} on two dialogue summarization benchmarks \cite{zhao2021todsum,yuan2019abstractive} demonstrate the effectiveness of our method.

In summary, our main contributions are:
\begin{itemize}[itemsep=0.3pt]
    \item We focus on improving the prompt transfer in prompt tuning from dialogue state tracking to few-shot dialogue summarization. To the best of our knowledge, SAPT is the first effective dialogue-specific prompt transfer technique.
    \item By training the model to perform skeleton generation during prompt transfer, SAPT provides extra supervision that essentially functions as an intermediate task-specific medium between the distinct source and target task, allowing the model to better consume the dialogue state information from the source task.
    \item To preserve model capability during prompt transfer, we design a novel approach that employs perturbation-based probes to automatically extract dialogue skeletons as supervised training data for skeleton generation, requiring neither annotation effort nor domain knowledge.
\end{itemize}

\begin{figure*}[ht]
\centering
\includegraphics[width=0.95\linewidth]{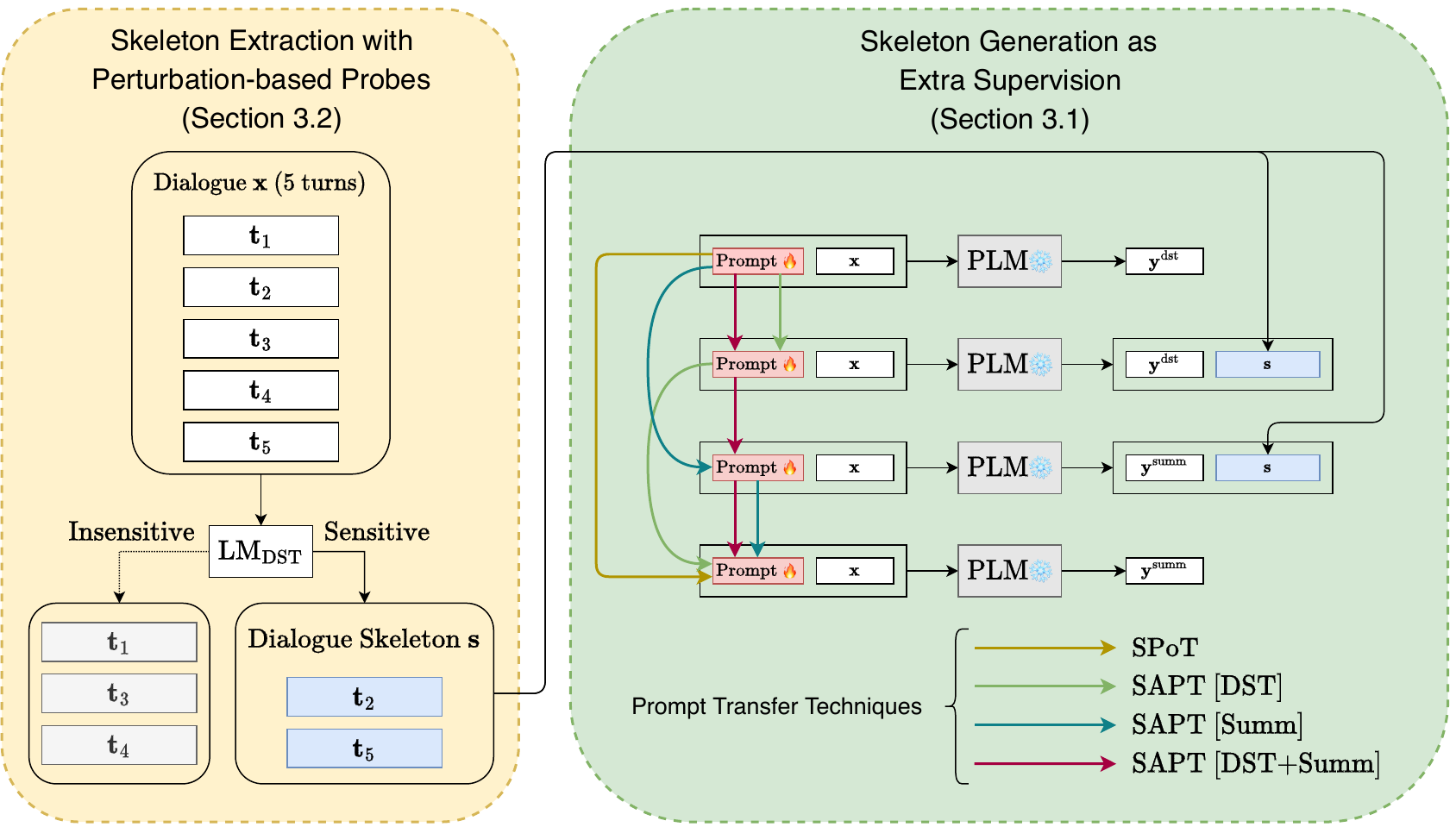}
\caption{The overall workflow of Skeleton-Assisted Prompt Transfer (SAPT). Besides the original task supervision ($\mathbf{y}^{\text{dst}}$ or $\mathbf{y}^{\text{summ}}$), SAPT uses skeleton generation as extra supervision (\S \ref{subsec:method-skeleton-generation}) by training on the dialogue skeletons $\mathbf{s}$ extracted with perturbation-based probes (\S \ref{subsec:method-probing-induced-skeleton}).}
\label{fig:model}
\end{figure*}

\section{Preliminaries}
\label{sec:preliminaries}

\subsection{Problem Definition}
\label{subsec:pre-problem-definition}

Abstractive dialogue summarization is typically formulated as a sequence-to-sequence problem \cite{nallapati-etal-2016-abstractive}. Given a dialogue history $\mathbf{x}$, a transformer-based encoder-decoder pretrained language model (PLM), $p_\theta(\mathbf{y}^{\text{summ}} | \mathbf{x})$, is trained to generate a summary $\mathbf{y}^{\text{summ}}$, where $\theta$ denotes the trainable parameters of the PLM. In this paper, we specifically study the dialogue summarization task in the few-shot setting \cite{brazinskas-etal-2020-shot}, meaning that there are only a limited number of annotated samples available for model training.

To mitigate the data scarcity problem, it is common to turn to transfer learning by leveraging massive supervised data from other related domains or tasks that could potentially provide useful knowledge. Dialogue state tracking (DST), a related task to dialogue summarization, aims to correctly infer the speaker's goal in the form of semantic slot-value pairs (\texttt{[slot, value]}) as a dialogue progresses, such as \texttt{[food, Italian]} and \texttt{[pricerange, high]}. We thus notice that the supervised data of the DST task should be able to bring conducive knowledge for the dialogue summarization task, as the semantic slots and values tracked by DST are expected to be covered in the dialogue summary. Besides the notable relevance between those two tasks, with DST being a language understanding task as opposed to dialogue summarization being a language generation task, the annotations of DST should arguably be easier to get in practice, compared to those of dialogue summarization. Therefore, we herein focus on how to perform effective transfer learning with ample supervised DST data to benefit the few-shot dialogue summarization.

Although the DST task is traditionally formulated as a classification problem, recent work \cite{lin-etal-2021-leveraging,zhao-etal-2021-effective-sequence} has shown the possibility of achieving competitive DST performance by treating DST as a sequence-to-sequence generation task. Specifically, conditioned on the dialogue history $\mathbf{x}$, the encoder-decoder model is trained to generate a sequence of tokens, in the format of \texttt{``slot1 is value1, slot2 is value2, ...''}, denoted as $\mathbf{y}^{\text{dst}}$. We thereby adopt this formulation for DST throughout our work so as to allow the generative encoder-decoder model's knowledge transfer (from DST to dialogue summarization) to happen. With the unified generative sequence-to-sequence-based DST and dialogue summarization, the conditional generation task can be formulated as follows ($\mathbf{y}$ can be either $\mathbf{y}^{\text{summ}}$ or $\mathbf{y}^{\text{dst}}$):
\begin{equation*}
P(\mathbf{y} | \mathbf{x}) = \prod_{i=1}^{|\mathbf{y}|} p_\theta(y_i | \mathbf{x}, y_{<i}).
\end{equation*}

\subsection{Prompt Transfer in Prompt Tuning}
\label{subsec:pre-prompt-tuning}

Among recent transfer learning techniques, prompt transfer \cite{vu-etal-2022-spot} in prompt tuning \cite{li-liang-2021-prefix,lester-etal-2021-power} has gained great popularity because of its parameter efficiency. We thus adopt it as our starting point for transfer learning from DST to dialogue summarization, and further improve it in \autoref{sec:method}.

Prompt tuning is a new paradigm of utilizing PLMs for downstream tasks. It is motivated by the intuition that PLMs can be steered with a proper context, without the need for any model parameter updates. In prompt tuning, a sequence of continuous trainable embeddings called ``soft prompt'', parameterized by $\phi$, is prepended to the input sequence. During training, all parameters of the PLM ($\theta$) are frozen, but unlike prompt design \cite{NEURIPS2020_1457c0d6} which searches for actual tokens in the discrete space, prompt tuning optimizes the ``soft prompt'' ($\phi$) directly in the continuous space, allowing it to be more expressive. The log-likelihood training objective can be formulated as follows:
\begin{equation*}
\label{eq:log-likelihood}
\max_\phi \log p_{\theta, \phi}(\mathbf{y} | \mathbf{x}) = \sum_{i=1}^{|\mathbf{y}|} \log p_{\theta, \phi}(y_i | \mathbf{x}, y_{<i}).
\end{equation*}

Prompt transfer is a simple yet effective transfer learning technique designed for prompt tuning. The soft prompt is first trained in the source task and then used as parameter initialization for the prompt tuning in the target task. Prompt transfer inherits the advantage of prompt tuning in terms of parameter efficiency, as its transfer learning process likewise relies merely on the lightweight soft prompt. \citet{su-etal-2022-transferability} show that prompt transfer generally works well in the transfer learning between language understanding tasks while it can only provide relatively mediocre performance in language generation tasks. This indicates the necessity to design task-specific prompt transfer approaches in language generation tasks such as dialogue summarization, which is exactly the central problem we focus on in this paper (detailed in \autoref{sec:method}).

\begin{algorithm}[ht] 
    \caption{Skeleton Extraction with Perturbation-based Probes}
    \label{alg:skeleton-extraction} 
    \begin{algorithmic}[1]
        \Require{}
        \Statex{a collection of dialogues $\mathcal{X}$ containing $N$ dialogues: $\mathcal{X} = \{\mathbf{x}_1,\mathbf{x}_2,\dots,\mathbf{x}_N\}$, where dialogue $\mathbf{x}_i$ contains $p_i$ dialogue turns: $\mathbf{x}_i = [\mathbf{t}_{i1},\mathbf{t}_{i2},\dots,\mathbf{t}_{i{p_i}}]$, $1\leq i \leq N$;}
        \Statex{a trained DST model $\text{LM}_{\text{DST}}$;}
        \Statex{a textual similarity metric $\text{\texttt{Sim}}(\cdot,\cdot)$ (higher means more similar).}
        \Ensure{}
        \Statex{a collection of dialogue skeletons: $\mathcal{S} = \{\mathbf{s}_1,\mathbf{s}_2,\dots,\mathbf{s}_N\}$, a subset $\text{\texttt{set}}(\mathbf{s}_i) \subseteq \text{\texttt{set}}(\mathbf{x}_i)$ for each dialogue $\mathbf{x}_i \in \mathcal{X}$, $1\leq i \leq N$}.
        \State{$\mathcal{M} = \{\}$}
        \For{$i=1,2,\dots,N$}
            \State{$\mathbf{o}_i = \text{LM}_{\text{DST}}(\mathbf{x}_i)$}
            \For{$j=1,2,\dots,p_i$}
                \State{$\mathbf{o}_{ij} = \text{LM}_{\text{DST}}(\mathbf{x}_i \setminus [\mathbf{t}_{ij}])$}
                \State{$m_{ij} = \text{\texttt{Sim}}(\mathbf{o}_i, \mathbf{o}_{ij})$}
                \State{add $m_{ij}$ to $\mathcal{M}$}
            \EndFor
        \EndFor
        \State{$\mathcal{S} = \{\}$}
        \State{$m_{\text{median}} = \text{\texttt{Median}}(\mathcal{M})$}
        \For{$i=1,2,\dots,N$}
            \State{$\mathbf{s}_i = [~]$}
            \For{$j=1,2,\dots,p_i$}
                \If{$m_{ij} < m_{\text{median}}$}
                    \State{append $\mathbf{t}_{ij}$ to $\mathbf{s}_i$}
                \EndIf
            \EndFor
            \State{add $\mathbf{s}_i$ to $\mathcal{S}$}
        \EndFor
        \State{\textbf{return }$\mathcal{S}$}
    \end{algorithmic}  
\end{algorithm}

\section{Method: Skeleton-Assisted Prompt Transfer (SAPT)}
\label{sec:method}

The existing non-task-specific general-purpose prompt transfer technique \cite{vu-etal-2022-spot} relies solely on the source and target task supervision to train the soft prompt, without the help of any intermediate task-specific medium. Even though DST and dialogue summarization are closely related tasks, the intrinsic domain shift between them should still not be ignored. Therefore, having an intermediate task-specific medium should conceivably be helpful for better connecting the distinct source and target task. Such a medium can take the form of extra task supervision separately incorporated into both the source and target task supervision, since in this way the updated source and target task have more overlap and get semantically closer to each other.

Also, as the model capability of processing source task data is closely associated with the knowledge it has gained during the source task pretraining, it needs to be effectively preserved during the prompt transfer to facilitate the target task. Nonetheless, the capability per se is admittedly a bit abstract and thus hard to concretely model in practice. Inspired by recent advances in interpretable NLP, we argue that the model sensitivity to perturbation-based probes should arguably be a concretization of model capability~\cite{talmor-etal-2020-olmpics}. Thus, maintaining model sensitivity during the prompt transfer should logically benefit the preservation of model capability. And notably, the aforementioned extra task supervision can exactly create conditions for (source-task) model-sensitivity information to be explicitly passed to the (target-task) model during the prompt transfer.

To these ends, we propose Skeleton-Assisted Prompt Transfer (SAPT), a dialogue-specific prompt transfer technique. SAPT provides the model with extra supervision during its prompt transfer by training it to perform skeleton generation along the way (detailed in \autoref{subsec:method-skeleton-generation}). This extra supervision (i.e. skeleton generation) is separately incorporated into both the source and target task supervision, and thus can essentially function as an \emph{intermediate task-specific medium} (because of the increased overlap between the updated source and target task) that is beneficial for the cross-task knowledge transfer.

To get the supervised training data for skeleton generation, we design a novel automatic skeleton extraction approach that requires neither annotation effort nor domain knowledge (detailed in \autoref{subsec:method-probing-induced-skeleton}). Specifically, we observe the model's output variation to perturbation-based probes and extract the dialogue turns to which the model displays the highest sensitivity as skeletons. Training the model on such skeletons can also help \emph{preserve model capability during prompt transfer}. This is because those skeletons (extracted with perturbation-based probes) embody the model sensitivity to perturbation-based probes which is a concretization of model capability.

On the whole, SAPT creates an intermediate task-specific medium using skeleton generation as extra supervision (\S \ref{subsec:method-skeleton-generation}), and preserves model capability during prompt transfer by training the model on the skeletons extracted with perturbation-based probes (\S \ref{subsec:method-probing-induced-skeleton}). As a result, the distinct source and target task is able to be better connected because they have got semantically closer to each other, and the target task is able to be facilitated because the model has been discouraged from forgetting the knowledge it has gained during the source task pretraining. \S \ref{subsec:workflow} describes SAPT's overall workflow.

\subsection{Skeleton Generation as Extra Supervision}
\label{subsec:method-skeleton-generation}

In SAPT, the skeleton generation task is incorporated into the original task (either the source or the target task, or both) as extra supervision. We denote a supervised sample of the original task as ($\mathbf{x}$, $\mathbf{y}$), where $\mathbf{x}$ represents the dialogue history and $\mathbf{y}$ represents the original task supervision that could be either the sequence-to-sequence-based dialogue state ground-truth or the dialogue summary ground-truth. For each sample ($\mathbf{x}$, $\mathbf{y}$), We also have a dialogue skeleton, denoted as $\mathbf{s}$, extracted from the dialogue history $\mathbf{x}$ (the skeleton extraction algorithm is detailed in \autoref{subsec:method-probing-induced-skeleton}). Such a dialogue skeleton is essentially an ordered collection of dialogue turns. For instance, if a dialogue history $\mathbf{x}$ contains $p$ dialogue turns, i.e. $\mathbf{x} = [\mathbf{t}_1, \mathbf{t}_2, \ldots, \mathbf{t}_p]$, its dialogue skeleton $\mathbf{s}$ will contain $q$ dialogue turns ($q \leq p$), denoted as $\mathbf{s} = [\mathbf{t}^s_1, \mathbf{t}^s_2, \ldots, \mathbf{t}^s_q]$, and thus $\texttt{set}(\mathbf{s}) \subseteq \texttt{set}(\mathbf{x})$. The dialogue skeleton $\textbf{s}$ is appended to the original task supervision $\textbf{y}$ as extra supervision, and the model is trained to perform the original task and then skeleton generation. The new log-likelihood training objective is:
\begin{align*}
&\max_\phi \log p_{\theta, \phi}(\mathbf{y} \oplus \mathbf{s} ~|~ \mathbf{x}) \\
&= \log p_{\theta, \phi}(\mathbf{y} | \mathbf{x}) + \log p_{\theta, \phi}(\mathbf{s} | \mathbf{x}, \mathbf{y}) \\
&= \log p_{\theta, \phi}(\mathbf{y} | \mathbf{x}) + \sum_{i=1}^{q} \log p_{\theta, \phi}(\mathbf{t}^s_i | \mathbf{x}, \mathbf{y}, \mathbf{t}^s_{<i}).\\
\end{align*}

\subsection{Skeleton Extraction with Perturbation-based Probes}
\label{subsec:method-probing-induced-skeleton}

We extract dialogue skeletons (used as supervised training data for skeleton generation in \autoref{subsec:method-skeleton-generation}) with perturbation-based probes. Given a dialogue in a collection of dialogues, $\mathbf{x}_i \in \mathcal{X}$, we first construct the perturbation-based probes by deleting a dialogue turn from $\mathbf{x}_i$ at a time. The resultant perturbation-based probes can be expressed as $\mathbf{x}_i \setminus [\mathbf{t}_{ij}], 1\leq j\leq p_i$ ($\mathbf{x}_i$ contains $p_i$ dialogue turns). We then feed those perturbation-based probes individually into the trained source-task (DST) model, $\text{LM}_{\text{DST}}$, and get the model output $\mathbf{o}_{ij}$ corresponding to each deleted dialogue turn $\mathbf{t}_{ij}$. In the meantime, we also feed the whole dialogue history $\mathbf{x}_i$ into $\text{LM}_{\text{DST}}$ and get the model output $\mathbf{o}_i$. Next, we compute the textual similarity score $m_{ij}$ between $\mathbf{o}_i$ and $\mathbf{o}_{ij}$ using a textual similarity metric $\text{\texttt{Sim}}(\cdot,\cdot)$ (higher means more similar). We execute the aforementioned procedure for each dialogue in $\mathcal{X}$. After that, we group together all the similarity scores we compute along the way and find the median of them. Finally, we extract those dialogue turns, whose corresponding similarity scores are less than the median, as the dialogue skeletons. Algorithm \ref{alg:skeleton-extraction} presents the process of extracting a dialogue skeleton $\mathbf{s}_i$ for each dialogue $\mathbf{x}_i \in \mathcal{X}$.

\subsection{Overall Workflow}
\label{subsec:workflow}

Built on top of \textsc{SPoT} \cite{vu-etal-2022-spot} while following the new training objective derived in \autoref{subsec:method-skeleton-generation}, SAPT uses skeleton generation as extra supervision by training on the dialogue skeletons extracted in \autoref{subsec:method-probing-induced-skeleton}. Skeleton generation (as extra supervision) can be separately incorporated into either the source (DST) or the target (dialogue summarization) task supervision, or both. We thereby propose three SAPT variants: \textsc{SAPT [DST]}, \textsc{SAPT [Summ]}, and \textsc{SAPT [DST+Summ]}.

As depicted in \autoref{fig:model}, \textsc{SAPT [DST+Summ]} includes four steps:
\begin{enumerate}[noitemsep]
    \item perform prompt tuning on the DST (source task) supervision;\label{step:1}
    \item perform prompt transfer from the previous step, and then perform prompt tuning on the DST (source task) \& skeleton generation supervision;\label{step:2}
    \item perform prompt transfer from the previous step, and then perform prompt tuning on the (few-shot) dialogue summarization (target task) \& skeleton generation supervision;\label{step:3}
    \item perform prompt transfer from the previous step, and then perform prompt tuning on the (few-shot) dialogue summarization (target task) supervision.\label{step:4}
\end{enumerate}

Compared to \textsc{SAPT [DST+Summ]}, \textsc{SAPT [DST]} omits step \#\ref{step:3} while \textsc{SAPT [Summ]} omits step \#\ref{step:2}; \textsc{SPoT} \cite{vu-etal-2022-spot} omits both step \#\ref{step:2} and step \#\ref{step:3}.

\begin{table*}[!ht]
\centering
\small
\begin{tabular}{l|ccc|ccc}
    \toprule
    \multicolumn{1}{c}{} & \multicolumn{3}{c}{\textsc{TODSum}} & \multicolumn{3}{c}{\textsc{SPNet}} \\
    \cmidrule(lr){2-4} \cmidrule(lr){5-7}
    Models & R-1 & R-2 & R-L & R-1 & R-2 & R-L \\ 
    \midrule
        \textsc{Prompt Tuning} \cite{lester-etal-2021-power} & 18.67 & 2.85 & 13.33 & 33.29 & 11.24 & 19.32 \\
        \textsc{SPoT} \cite{vu-etal-2022-spot} & 56.96 & 30.26 & 38.40 & 45.46 & 33.27 & 39.49 \\
    \midrule
        \textsc{SAPT [DST]} & 62.00 & 36.95 & 43.13 & 53.43 & 40.07 & 44.92 \\
        \textsc{SAPT [Summ]} & 57.39 & 34.60 & 42.50 & 49.65 & 37.30 & 42.57 \\
        \textsc{SAPT [DST+Summ]} & \textbf{62.25} & \textbf{40.75} & \textbf{48.30} & \textbf{56.49} & \textbf{41.93} & \textbf{47.46} \\
    \bottomrule
\end{tabular}
\caption{Few-shot (100-shot) results on the full \textsc{TODSum} \cite{zhao2021todsum} and \textsc{SPNet} \cite{yuan2019abstractive} test set. All three \textsc{SAPT} variants outperform the baseline model on both datasets, \textsc{SPoT} \cite{vu-etal-2022-spot}. \textsc{SAPT [DST+Summ]} achieves the highest ROUGE scores with \textbf{significant} performance improvements.}
\label{table:res}
\end{table*}

\begin{table*}[!ht]
\centering
\small
\begin{tabular}{l|cccc}
    \toprule
     & \textbf{Informativeness} & \textbf{Faithfulness} & \textbf{Fluency} & \textbf{Redundancy} \\ 
    \midrule
        Ground Truth & 1.92 & 1.90 & 1.95 & 1.97 \\
    \midrule
        \textsc{SPoT} \cite{vu-etal-2022-spot} & 1.77 & 1.70 & 1.73 & 1.71 \\
    \midrule
        \textsc{SAPT [DST]} & 1.82 & 1.77 & 1.85 & 1.79 \\
        \textsc{SAPT [Summ]} & 1.80 & 1.76 & 1.85 & 1.73 \\
        \textsc{SAPT [DST+Summ]} & \textbf{1.86} & \textbf{1.82} & \textbf{1.90} & \textbf{1.81} \\
    \bottomrule
\end{tabular}
\caption{Human evaluation results in terms of the informativeness, faithfulness, fluency, and redundancy of the generated summaries on \textsc{TODSum} test set. \textsc{SAPT [DST+Summ]} consistently performs the best across all metrics.}
\label{table:human-eval}
\end{table*}

\begin{table*}[!ht]
\centering
\small
\resizebox{\textwidth}{!}{
\begin{tabular}{l|ccc|l|ccc|l|ccc}
    \toprule
    \textbf{skeleton type} & R-1 & R-2 & R-L & \textbf{decoding order} & R-1 & R-2 & R-L & \textbf{source \& target task supervision} & R-1 & R-2 & R-L \\ 
    \midrule
        \multicolumn{12}{c}{\textsc{SAPT [DST]}} \\
    \midrule
        random skeleton & 57.89 & 37.04 & 42.47 & prepended skeleton & 58.85 & 34.17 & 42.80 & w/o source task supervision & 59.61 & 32.26 & 41.81 \\
        our skeleton & 62.00 & 36.95 & 43.13 & appended skeleton & 62.00 & 36.95 & 43.13 & w/ source task supervision & 62.00 & 36.95 & 43.13 \\
    \midrule
        \multicolumn{12}{c}{\textsc{SAPT [Summ]}} \\
    \midrule
        random skeleton & 55.14 & 33.07 & 41.70 & prepended skeleton & 53.33 & 31.16 & 39.34 & w/o target task supervision & 57.81 & 32.03 & 41.02 \\
        our skeleton & 57.39 & 34.60 & 42.50 & appended skeleton & 57.39 & 34.60 & 42.50 & w/ target task supervision & 57.39 & 34.60 & 42.50 \\
    \midrule
        \multicolumn{12}{c}{\textsc{SAPT [DST+Summ]}} \\
    \midrule
        random skeleton & 58.34 & 38.98 & 43.11 & prepended skeleton & 62.04 & 40.92 & 46.74 & w/o source \& target task supervision & 59.85 & 32.46 & 41.62 \\
        our skeleton & 62.25 & 40.75 & 48.30 & appended skeleton & 62.25 & 40.75 & 48.30 & w/ source \& target task supervision & 62.25 & 40.75 & 48.30 \\
    \bottomrule
\end{tabular}
}
\caption{Results of ablation studies on the effect of skeleton type, decoding order, and source \& target task supervision for all three \textsc{SAPT} variants on \textsc{TODSum} test set.}
\label{table:ablation}
\end{table*}

\section{Experiment}
\label{sec:experiments}

\subsection{Dataset and Baseline}
\label{subsec:dataset-and-model}

To study the cross-task prompt transfer from dialogue state tracking (DST) to few-shot dialogue summarization, we perform experiments on a DST dataset: MultiWOZ 2.2 \cite{zang-etal-2020-multiwoz}, and on two task-oriented dialogue summarization datasets: \textsc{TODSum} \cite{zhao2021todsum} and \textsc{SPNet} \cite{yuan2019abstractive}. MultiWOZ 2.2 is an error-fixed version of MultiWOZ \cite{budzianowski-etal-2018-multiwoz}, which is a classic task-oriented multi-domain dialogue dataset containing over 10,000 annotated dialogues and has been extensively used for studying DST.  \textsc{TODSum} and \textsc{SPNet} are both constructed using the dialogues from MultiWOZ, and differ mainly in terms of summary style and length. On average, the summaries in \textsc{SPNet} are roughly two times longer than those in \textsc{TODSum} (96.4 vs. 45.4 words). To evaluate our method under the few-shot setting, on each dialogue summarization dataset we randomly choose 100 samples from the training set for model training and test on the full test set.

We use BART-large\footnote{\url{https://huggingface.co/docs/transformers/model\_doc/bart}} \cite{lewis-etal-2020-bart} as the backbone throughout the experiments. We focus on the comparison between prompt-tuning-based methods, as they have been proven to be able to maintain as comparable performance as the adapter-based methods while being much more parameter-efficient \cite{li-liang-2021-prefix,vu-etal-2022-spot}. We choose \textsc{SPoT} \cite{vu-etal-2022-spot} as the baseline method, which has been commonly used as a parameter-efficient transfer learning technique. \autoref{sec:appendix} presents the implementation details.

\subsection{Automatic Evaluation}
\label{subsec:automatic-evaluation}

We use the widely-used ROUGE metrics \cite{lin-2004-rouge} as automatic evaluation metrics, including ROUGE-1 (R-1), ROUGE-2 (R-2), and ROUGE-L (R-L) F1 scores with rouge-score python package\footnote{\url{https://pypi.org/project/rouge-score/}}. Few-shot (100-shot) results are presented in \autoref{table:res}, where we also attach the results of \textsc{Prompt Tuning} \cite{li-liang-2021-prefix}. Unsurprisingly, \textsc{Prompt Tuning} performs badly without any knowledge transfer, which indicates the necessity of conducting prompt transfer from DST to few-shot dialogue summarization. Among different prompt transfer techniques, all three \textsc{SAPT} variants outperform the baseline method \textsc{SPoT} on both datasets, suggesting the effectiveness of the proposed \textsc{SAPT} method. It is also observed that \textsc{SAPT[DST]} consistently outperforms \textsc{SAPT[Summ]}. We attribute this to the fact that there are much more dialogue samples (along with their extracted dialogue skeletons) that are used for the preservation of model capability during step \#\ref{step:2} than step \#\ref{step:3}, as in step \#\ref{step:3} we only make use of 100 dialogue samples dedicated for few-shot dialogue summarization. Notably, when both step \#\ref{step:2} and step \#\ref{step:3} are executed, \textsc{SAPT[DST+Summ]} is able to further improve the performance by a significant margin, compared to \textsc{SAPT[DST]} and \textsc{SAPT[Summ]}. This demonstrates the effectiveness of creating an intermediate task-specific medium between the source DST task and the target few-shot dialogue summarization task (by incorporating the skeleton generation task into both of them).

\subsection{Human Evaluation}
\label{subsec:human-evaluation}

To further evaluate the generated summaries, we perform a human evaluation via crowdsourcing. We randomly select 100 samples from \textsc{TODSum} test set and run different models on them to generate summaries. We recruit human participants on Prolific\footnote{\url{https://www.prolific.co/}}, a crowdsourcing platform, to rate the generated summaries (and also the ground-truth summaries) from 0 to 2 in terms of four evaluation metrics: informativeness, faithfulness, fluency, and redundancy\footnote{Details of the metrics can be found in \autoref{sec:human-eval-metrics}.}. Each summary instance is evaluated by 5 different human participants, and the inter-annotator agreement (IAA) score for each metric is 0.577, 0.635, 0.649, 0.591, with an average IAA of 0.613. Results shown by the average scores in \autoref{table:human-eval} are consistent with the automatic evaluation results: all three \textsc{SAPT} variants outperform the baseline method \textsc{SPoT}, and \textsc{SAPT[DST+Summ]} consistently performs the best across all metrics. Meanwhile, all generated summaries are deemed to be worse than the ground-truth summaries, meaning that there is still room for these summarization models to be improved. We also conduct a case study by ourselves, detailed in \autoref{sec:case-study}.

\subsection{Ablation Study}
\label{subsec:analysis}

To fully investigate the effectiveness of \textsc{SAPT}, we study the impact of skeleton type, decoding order, and source \& target task supervision. \autoref{table:ablation} shows the results of ablation studies.

\noindent
\textbf{Skeleton Type. } We replace the extracted skeletons (\S \ref{subsec:method-probing-induced-skeleton}) with randomly-extracted skeletons. We make sure that in total half of the dialogue turns are selected as skeletons to align with our usage of \texttt{Median()}, and that there is at least one dialogue turn selected for each dialogue. The observed performance drop demonstrates the effectiveness of our skeletons extracted with perturbation-based probes. Models with random skeleton still outperform \textsc{SPoT} in general, and we attribute this to the possible match between random skeleton and our skeleton, and also the imperfect intermediate task-specific medium which persists in the workflow.

\noindent
\textbf{Decoding Order. } We prepend the skeletons instead of appending them. The observed performance drop demonstrates that the original task supervision needs to be prioritized, and prepending makes it more difficult for models to learn the cross-task knowledge.

\noindent
\textbf{Source \& Target Task Supervision. } We remove all the original task supervision along the way. The observed performance drop is as expected, but the superior performance against \textsc{SPoT} demonstrates the benefit our skeletons bring for cross-task knowledge transfer.

\section{Related Work}
\label{sec:related-work}

\textbf{Parameter-Efficient Transfer Learning. }
To efficiently make use of pretrained language models (PLMs) \cite{devlin-etal-2019-bert,JMLR:v21:20-074,lewis-etal-2020-bart,NEURIPS2020_1457c0d6}, \citet{li-liang-2021-prefix} propose to prepend continuous trainable task-specific embeddings to the input sequence while keeping the entire PLM frozen. \citet{lester-etal-2021-power} provide a simplified approach, named prompt tuning, which becomes more competitive with model fine-tuning as scale increases. To enable cross-task knowledge transfer \cite{ruder2017overview,liu-etal-2019-multi} under the prompt tuning paradigm, \citet{vu-etal-2022-spot} propose SPoT, which learns soft prompts from source tasks as initialization for target tasks.
\citet{su-etal-2022-transferability} further explore the transferability of soft prompts across different downstream tasks.
Built on top of \citet{vu-etal-2022-spot}, our method is able to improve the effectiveness of cross-task prompt transfer in few-shot dialogue summarization.

\noindent
\textbf{Low-Resource Abstractive Summarization. }
Multiple lines of approaches have been proposed to mitigate the data scarcity problem in abstractive summarization, such as reinforcement learning \cite{kohita-etal-2020-q,hyun-etal-2022-generating}, self-supervised learning \cite{fu-etal-2021-repsum,wang-wan-2021-transsum,zhuang-etal-2022-learning}, data augmentation \cite{amplayo-lapata-2020-unsupervised,laskar-etal-2020-wsl,fabbri-etal-2021-improving,chen-yang-2021-simple}, model pretraining or fine-tuning with in-domain unlabeled data or out-of-domain labeled data \cite{yang-etal-2020-ted,goodwin-etal-2020-towards,yu-etal-2021-adaptsum,zou-etal-2021-low,magooda-etal-2021-exploring-multitask}, and few-shot learning via adapters \cite{brazinskas-etal-2020-shot,brazinskas-etal-2022-efficient} or prompt tuning \cite{zhao-etal-2022-domain,liu-etal-2022-psp,yuan-etal-2022-shot}. In this paper, we focus on the few-shot dialogue summarization and improve it by ameliorating cross-task prompt transfer in prompt tuning with cross-task labeled data.

\noindent
\textbf{Perturbation-based Probes. }
In interpretable NLP, while probes sometimes refer to algorithms or models aiming to extract information from continuous embeddings \cite{adi2017finegrained}, they can also refer to textual inputs designed for acquiring model outputs that are either useful for downstream tasks \cite{petroni-etal-2019-language,zhong-etal-2021-factual} or informative for model interpretability \cite{goldberg2019assessing,bacon2019does,xie-etal-2022-calibrating}. Perturbation-based probes, which fall into the latter category, have gained popularity because of their simplicity and cost-efficiency. For instance, \citet{sankar-etal-2019-neural,abdou-etal-2020-sensitivity,ettinger-2020-bert,clouatre-etal-2022-detecting} investigate the sensitivity of neural language models to input perturbation; \citet{richardson-sabharwal-2020-qa,talmor-etal-2020-olmpics,bitton-etal-2021-automatic,gupta-etal-2022-model} utilize perturbation to construct better NLP testbeds.
In contrast, we leverage perturbation-based probes to automatically extract skeletons from dialogues.

\section{Conclusion}
\label{sec:conclusion}

We focus on improving the prompt transfer in prompt tuning from dialogue state tracking to few-shot dialogue summarization, and propose \textsc{SAPT}, a dialogue-specific prompt transfer technique, which uses skeleton generation as extra supervision by training the model on the dialogue skeletons extracted with perturbation-based probes. In this way, a beneficial intermediate task-specific medium is created between the source and target task, and the model capability is able to be better preserved during the prompt transfer, resulting in the model's better consumption of dialogue state information from the source task. Significantly stronger empirical performance and in-depth analyses on two dialogue summarization benchmarks demonstrate the effectiveness of our method in few-shot dialogue summarization.

\section{Limitations}
\label{sec:limitations}

Despite the strong performance achieved by SAPT, we use the pre-trained language model (PLM) as the backbone of our method. Therefore, we cannot go beyond the limitation of the maximum sequence length of the PLM. In fact, long-form language understanding and generation have been widely acknowledged as an open research question that needs much further investigation, which is beyond the scope of our paper.

\section{Ethics \& Broader Impacts}
\label{sec:ethics-and-broader-impacts}

All datasets used in this work are public. We did not collect any personal information from our human participants nor did we present them with any harmful model outputs. Our dialogue summarization models face the same potential pitfalls as other contemporary language learning systems do, e.g. being prone to echoing the biases present in the dataset \cite{sheng-etal-2019-woman}.

\bibliography{anthology,custom}

\appendix

\section{Data Annotation Study}
\label{sec:data-annotation-appendix}

We recruit 30 human participants on Prolific\footnote{\url{https://www.prolific.co/}}, a crowdsourcing platform, to annotate 30 dialogues for their dialogue states and dialogue summaries. We split 30 participants into two batches and split 30 dialogues into two batches as well. We follow a Latin Square design, similarly to \cite{gonzalez2020reverse}, to make sure that each batch of participants only sees each batch of dialogues in one of the following two annotation settings: dialogue state and dialogue summary, yet each setting is tested on both all 30 annotators and all 30 dialogues. This ensures that no bias in the duration of annotation occurs due to annotators having previously seen the dialogues.

We measure the duration of the annotation processes for both dialogue state and dialogue summary. The average duration of annotating a dialogue for its dialogue states is 1.3 minutes; the average duration of annotating a dialogue for its dialogue summary is 3.8 minutes, which is much longer. These results are in line with our intuition: the annotation of a dialogue summary requires not only tracking the dialogue states, but also having an utterance-level detailed understanding of the dialogue, because only after understanding the whole dialogue progression can annotators write a fluent and faithful summary.

\section{Implementation Details}
\label{sec:appendix}

We use Hugging Face Transformers\footnote{\url{https://github.com/huggingface/transformers}} \cite{wolf-etal-2020-transformers} during implementation. We train the BART-large models using AdamW \cite{loshchilov2018decoupled} with the default learning rate linearly decaying from $5E-5$. All models with a prompt length of 200 are trained for 50 epochs on an NVIDIA TITAN Xp GPU (12 GB memory) with a batch size of 2 and they each take approximately 25 hours (for DST) / 0.3 hours (for 100-shot dialogue summarization) to train. During inference, we perform a beam search with a beam size of 6, and the decoding takes 1.5 seconds per batch.

All turns of the input dialogue are prepended with special tokens as speaker identifiers (\texttt{[USER]} or \texttt{[SYSTEM]}), and then concatenated into a single input sequence which is truncated to 1024 BPE tokens. We use the ROUGE-L F1 score as the textual similarity metric $\text{\texttt{Sim}}(\cdot,\cdot)$ in Algorithm \ref{alg:skeleton-extraction}. The dialogue skeletons are appended to the ground-truth dialogue states (or summaries), and there is a special token \texttt{[SEP]} between the dialogue states (or summaries) and skeletons.

\section{Details of Human Evaluation Metrics}
\label{sec:human-eval-metrics}

Human participants are asked to read the summaries and give their ratings (0, 1, or 2) in terms of four evaluation metrics:

\begin{itemize}
    \item \textbf{Informativeness} examines whether the critical information in the dialogue is missed in the summary:
        \begin{itemize}[label=$\star$]
            \item 0: lots of the critical information in the dialogue is missed;
            \item 1: a small amount of the critical information in the dialogue is missed;
            \item 2: no critical information in the dialogue is missed.
        \end{itemize}
    \item \textbf{Faithfulness} examines whether the information presented in the summary is factually incorrect or unmentioned according to the dialogue:
        \begin{itemize}[label=$\star$]
            \item 0: lots of the information presented in the summary is factually incorrect or unmentioned;
            \item 1: a small amount of the information presented in the summary is factually incorrect or unmentioned;
            \item 2: no information presented in the summary is factually incorrect or unmentioned.
        \end{itemize}
    \item \textbf{Fluency} examines whether the sentences in the summary are ungrammatical or ill-formed:
        \begin{itemize}[label=$\star$]
            \item 0: lots of the sentences in the summary are ungrammatical or ill-formed;
            \item 1: a small amount of the sentences in the summary are ungrammatical or ill-formed;
            \item 2: no sentence in the summary is ungrammatical or ill-formed.
        \end{itemize}
    \item \textbf{Redundancy} examines whether the expressions of the summary can be simplified:
        \begin{itemize}[label=$\star$]
            \item 0: lots of the expressions of the summary can be simplified;
            \item 1: a small amount of the expressions of the summary can be simplified;
            \item 2: no expression of the summary can be simplified.
        \end{itemize}
\end{itemize}

\section{Case Study}
\label{sec:case-study}

We present a case study in \autoref{tab:case-study} to illustrate the effectiveness of \textsc{SAPT}.

\begin{table*}[t!]
\centering
\small
\begin{tabular}{p{0.90\linewidth}}
\toprule
\textbf{Dialogue Example} \\
\midrule[0.03em]
\\

I am looking for a taxi from {\color{red} {yu garden}} arriving by {\color{blue} {14:30}}. 

\begin{FlushRight}
\textit{I'll need a destination to give them before I can book that for you.}
\end{FlushRight}

I'd be going to {\color{orange} {holy trinity church}}. I'll need the {\color{teal} {car type and contact number}} as well.

\begin{FlushRight}
\textit{The car that is reserved for you is a grey Toyota, and the contact number is 07997985529.  Can I do anything else for you?}
\end{FlushRight}

Nope that's all thank you.

\begin{FlushRight}
\textit{Alright, enjoy your trip!}
\end{FlushRight}

\\
\toprule[0.03em]
\textsc{SPoT} \cite{vu-etal-2022-spot} \\
\midrule[0.03em]
The user asks the user to book a taxi from {\color{red} {yu garden}} arriving by {\color{blue} {14:30}}, and the destination is {\color{orange} {holy trinity church}}.\\

\toprule[0.03em]
\textsc{SAPT [DST]} \\
\midrule[0.03em]
The user asks for the {\color{teal} {car type and contact number}} of the taxi. The taxi leaves at {\color{blue} {14:30}} on thursday, and the destination is {\color{orange} {holy trinity church}}.\\

\toprule[0.03em]
\textsc{SAPT [Summ]} \\
\midrule[0.03em]
The user wants to know the {\color{teal} {car type and the contact number}} of the taxi. The taxi leaves at {\color{blue} {14:30}}, and the destination is {\color{orange} {holy trinity church}}.\\

\toprule[0.03em]
\textsc{SAPT [DST+Summ]} \\
\midrule[0.03em]
The user asks the agent to check the {\color{teal} {car type and contact number}} of the taxi. The taxi leaves at {\color{blue} {14:30}} on thursday, from {\color{red} {yu garden}} to {\color{orange} {holy trinity church}}.\\

\toprule[0.03em]
Ground Truth \\
\midrule[0.03em]
The user wonders if it is possible to know the {\color{teal} {car type and the phone number}}. The taxi arrives at {\color{blue} {14:30}}, from {\color{red} {yu garden}} to {\color{orange} {holy trinity church}}.\\

\bottomrule
\end{tabular}
\caption{A case study. We highlight all dialogue-state-related information. The summaries provided by all three \textsc{SAPT} variants provide more complete dialogue-state-related information coverage than the baseline method \textsc{SPoT}. Among those three variants, only \textsc{SAPT [DST+Summ]} covers all dialogue-state-related information. However, compared to the ground truth, the summary provided by \textsc{SAPT [DST+Summ]} contains information that is unmentioned in the dialogue (i.e. on Thursday), suggesting there is still room for \textsc{SAPT} to be improved.}
\label{tab:case-study}
\end{table*}

\end{document}